\documentclass[11pt]{article}
\pdfoutput=1
\usepackage{acl2014}
\usepackage{times}
\usepackage{url}
\usepackage{latexsym}

\usepackage[USenglish]{babel}
\usepackage{lipsum}
\usepackage{amsmath}
\usepackage{amssymb}
\usepackage{booktabs}
\usepackage{multirow}
\usepackage{graphicx}
\usepackage{caption}
\usepackage{subcaption}
\usepackage{xspace}

\title{Learning Bilingual Word Representations by Marginalizing Alignments}
\author{Tom\'a\v{s} Ko\v{c}isk\'y \And
  Karl Moritz Hermann  \vspace{0.2cm}\\
  Department of Computer Science \\
  University of Oxford  \\
  Oxford, OX1 3QD, UK \\
  {\small\tt \{tomas.kocisky,karl.moritz.hermann,phil.blunsom\}@cs.ox.ac.uk} \\\And
  Phil Blunsom \\
}

\date{}

\newcommand{\prob}[1]{p\!\left(#1\right)}
\newcommand{\given}{|}

\newcommand{\svar}[2]{\mathbf{#1}}
\newcommand{\evar}[3]{{#1}_{#3}}

\newcommand{\ener}[1]{\mathrm{E}\!\lpa#1\rpa}
\newcommand{\lsq}{\left[}
\newcommand{\rsq}{\right]}
\newcommand{\lpa}{\left(}
\newcommand{\rpa}{\right)}
\newcommand{\fastalign}{{\sc FA}\xspace}
\newcommand{\adagrad}{{\sc AdaGrad}\xspace}
\newcommand{\BLEU}{{\sc BLEU}\xspace}
\newcommand{\AER}{{\sc AER}\xspace}
\newcommand{\fastalignLBL}{{\sc DWA}\xspace}
\newcommand{\model}{{\sc DWA}\xspace}
\usepackage[mathscr]{euscript}
\newcommand{\transp}{T}

\begin{document}
\maketitle
\begin{abstract}
  We present a probabilistic model that simultaneously learns alignments and
  distributed representations for bilingual data.
  By marginalizing over word alignments the model captures a larger semantic
  context than prior work relying on hard alignments.
  The advantage of this approach is demonstrated in a~cross-lingual
  classification task, where we outperform the prior published state of~the art.
\end{abstract}

\section{Introduction}

Distributed representations have become an increasingly important tool in
machine learning. Such representations---typically continuous vectors learned in
an unsupervised setting---can frequently be used in place of hand-crafted, and
thus expensive, features. By providing a richer representation than what can be
encoded in discrete settings, distributed representations have been
successfully used in many areas. This includes
AI and reinforcement learning~\cite{mnih-atari-2013}, image
retrieval~\cite{kirosmultimodal},
language modelling \cite{bengio2003neural}, sentiment analysis
\cite{Socher:2011:SRA:2145432.2145450,Hermann:2013}, frame-semantic parsing
\cite{Hermann:2014:ACLgoogle}, and document classification
\cite{klementiev-titov-bhattarai:2012:PAPERS}.

In Natural Language Processing (NLP), the use of distributed representations is
motivated by the idea that they could capture semantics and/or syntax, as well
as encoding a continuous notion of similarity, thereby enabling information
sharing between similar words and other units. The success of distributed
approaches to a number of tasks, such as listed above, supports this notion and
its implied benefits (see also \newcite{Turian:2010} and
  \newcite{Collobert:2008:UAN:1390156.1390177}).

While most work employing distributed representations has focused on monolingual
tasks, multilingual representations would also be useful for several NLP-related
tasks. Such problems include document classification, machine translation,
and cross-lingual information retrieval, where multilingual data is frequently the norm.
Furthermore, learning multilingual representations can also be useful for
cross-lingual information transfer, that is exploiting resource-fortunate
languages to generate supervised data in resource-poor ones.

We propose a probabilistic model that simultaneously learns word alignments and
bilingual distributed word representations. As opposed to previous work in this
field, which has relied on hard alignments or bilingual lexica
\cite{klementiev-titov-bhattarai:2012:PAPERS,DBLP:journals/corr/MikolovLS13}, we
marginalize out the alignments, thus capturing more bilingual semantic context. Further,
this results in our distributed word alignment (\model) model being the first
probabilistic account of bilingual word representations.
This is desirable as it allows better reasoning about the derived
representations and furthermore, makes the model suitable for inclusion in
higher-level tasks such as machine translation.

The contributions of this paper are as follows. We present a new probabilistic
similarity measure which is based on an alignment model and prior language modeling
work which learns and relates word representations across languages.
Subsequently, we apply these embeddings to a standard document classification task and show that they outperform the current published state of the art
\cite{Hermann:2014:ACLphil}.
As a by-product we develop a distributed version of \textsc{FastAlign}
\cite{naacl2013:dyer}, which performs on par with the original model, thereby
demonstrating the efficacy of the learned bilingual representations.

\section{Background}

The IBM alignment models, introduced by \newcite{Brown:1993:MSM:972470.972474},
form the basis of most statistical machine translation systems. In this paper we
base our alignment model on \textsc{FastAlign}~(\fastalign{}), a variation of
IBM model~2 introduced by \newcite{naacl2013:dyer}. This model is both fast and
produces alignments on par with the state of the art. Further, to induce the
distributed representations we incorporate ideas from the log-bilinear language
model presented by \newcite{Mnih:2007:TNG:1273496.1273577}.

\subsection{IBM Model~2}

Given a parallel corpus with aligned sentences, an alignment model can be used
to discover matching words and phrases across languages. Such models are an
integral part of most machine translation pipelines.  An alignment model learns
$\prob{\svar fs,\svar as \given \svar es}$ (or $\prob{\svar es,\svar {a'}s\given
\svar fs}$) for the source and target sentences $\svar es$ and $\svar fs$
(sequences of words). $\svar ax$ represents the word alignment across these two
sentences from source to target.
IBM model~2 \cite{Brown:1993:MSM:972470.972474} learns alignment and translation
probabilities in a generative style as follows:
\vspace{-0.5em}
\begin{equation*}
  \prob{\svar fs,\svar as \given \svar es}
    = \prob{J\given I}
      \prod_{j=1}^J \prob{\evar asj\given j,I,J} \prob{\evar fsj \given \evar es{a_j}},
\vspace{-0.5em}
    \end{equation*}
where $\prob{J\given I}$ captures the two sentence lengths;
$\prob{\evar asj\given j,I,J}$ the alignment and
$\prob{\evar fsj \given \evar es{a_j}}$ the translation probability.
Sentence likelihood is given by marginalizing out the alignments, which results
in the following equation:
\vspace{-0.5em}
\begin{equation*}
  \prob{\svar fs\given \svar es}
  = \prob{J\given I}
  \prod_{j=1}^J\sum_{i=0}^I
  \prob{i\given j,I,J} \prob{\evar fsj \given \evar esi} .
\vspace{-0.5em}
\end{equation*}
We use \textsc{FastAlign}~(\fastalign) \cite{naacl2013:dyer}, a log-linear
reparametrization of IBM model~2.
This model uses an alignment distribution defined by a single parameter
that measures how close the alignment is to the diagonal. This replaces the
original multinomial alignment distribution which often suffered from sparse
counts.  This improved model was shown to run an order of magnitude faster than
IBM model~4 and yet still outperformed it in terms of the \BLEU\ score and, on
Chinese-English data, in alignment error rate~(\AER).

\subsection{Log-Bilinear Language Model}

Language models assign a probability measure to sequences of words.
We use the log-bilinear language model proposed by
\newcite{Mnih:2007:TNG:1273496.1273577}. 
It is an n-gram based model defined in terms of an energy function
$\ener{w_n;w_{1:n-1}}$. The probability for predicting the next word $w_n$ given
its preceding context of $n-1$ words is expressed using the energy function
\begin{equation*}
  \ener{w_n;w_{1:n-1}}\!=\!\scalebox{0.75}[1.0]{\(-\)}\!\lpa
  \sum_{i=1}^{n-1}r_{w_i}^\transp C_i\rpa\!r_{w_n}
                        \scalebox{0.75}[1.0]{\(-\)}b_r^\transp r_{w_n}
                        \scalebox{0.75}[1.0]{\(-\)}b_{w_n} \label{eq:lbllm} \nonumber
\end{equation*}
as $
\prob{w_n\given w_{1:n-1}} = 
                      \frac1{Z_c}\exp\lpa-\ener{w_n;w_{1:n-1}}\rpa
                         \nonumber 
$ 
where
$Z_c =  \sum_{w_n}\exp\lpa-\ener{w_n;w_{1:n-1}}\rpa$ is the normalizer,
\mbox{$r_{w_i}\in\mathbb{R}^d$} are word representations,
\mbox{$C_i\in\mathbb{R}^{d\times d}$}~are context transformation matrices,
and \mbox{$b_r\in\mathbb{R}^d$}, \mbox{$b_{w_n}\in\mathbb{R}$} are
representation and word biases respectively.
Here, the sum of the transformed context-word vectors endeavors to be
close to the word we want to predict, since the likelihood in the model is
maximized when the energy of the observed data is minimized.

This model can be considered a variant of a log-linear language model in which,
instead of defining binary n-gram features, the model learns the features of the
input and output words, and a transformation between them.
This provides a
vastly more compact parameterization of a language model as 
n-gram features are not stored.

\subsection{Multilingual Representation Learning}

There is some recent prior work on multilingual distributed representation
learning.
Similar to the model presented here,
\newcite{klementiev-titov-bhattarai:2012:PAPERS} and \newcite{Zou:2013} learn
bilingual embeddings using word alignments. These two models are
non-probabilistic and conditioned on the output of a separate alignment model,
unlike our model, which defines a probability distribution over translations and
marginalizes over all alignments. These models are also highly
related to prior work on bilingual lexicon induction
\cite{haghighi-EtAl:2008:ACLMain}.
Other recent approaches include \newcite{Chandar:2013}, \newcite{Lauly:2013} and
\newcite[2014b]{Hermann:2014:ICLR}\nocite{Hermann:2014:ACLphil}. These models
avoid word alignment by transferring information across
languages using a composed sentence-level representation.

While all of these approaches are related to the model proposed in this paper,
it is important to note that our approach is novel by providing a probabilistic
account of these word embeddings. Further, we learn word alignments and
simultaneously use these alignments to guide the representation learning, which
could be advantageous particularly for rare tokens, where a sentence based
approach might fail to transfer information.

Related work also includes \newcite{DBLP:journals/corr/MikolovLS13}, who learn a
transformation matrix to reconcile monolingual embedding spaces, in an $l_2$~norm sense,
using dictionary entries instead of alignments, as well as
\newcite{Schwenk:2007} and \newcite{Schwenk:2012}, who also use distributed
representations for estimating translation probabilities.
\newcite{Faruqui:2014:EACL} use a technique based on CCA and alignments to
project monolingual word representations to a common vector space.

\section{Model}

Here we describe our distributed word alignment (\model) model.
The \model model can be viewed as a distributed extension of the
\fastalign model in that it uses a similarity measure over distributed word
representations instead of the standard multinomial translation probability
employed by~\fastalign.
We do this using a modified version of the log-bilinear language model in place
of the translation probabilities $\prob{\evar fsj\given \evar esi}$ at the heart
of the \fastalign model. This allows us to learn word representations for
both languages, a translation matrix relating these vector spaces, as well as alignments at the same time.

Our modifications to the log-bilinear model are as follows.  Where the original
log-bilinear language model uses context words to predict the next word---this
is simply the distributed extension of an n-gram language model---we use a word
from the source language in a parallel sentence to predict a target word.
An additional aspect of our model, which demonstrates its flexibility, is that
it is simple to include further context from the source sentence, such as words
around the aligned word or syntactic and semantic annotations. In this paper we
experiment with a transformed sum over $k$~context words to each side of the
aligned source word.
We evaluate different context sizes and report the results in Section~\ref{sec:experiments}.
We define the energy function for the translation probabilities to be
\vspace{-0.5em}
\begin{equation}
  \ener{f,e_i} = -\lpa\sum_{s=-k}^k r_{e_{i+s}}^\transp T_s\rpa r_f-b_r^\transp r_f-b_f  \label{eq:energy}
\vspace{-0.5em}
\end{equation}
where $r_{e_i},r_f\in\mathbb{R}^d$ are vector representations for source and target words
$e_{i+s}\in V_E,f\in V_F$ in their respective vocabularies,
$T_s\in\mathbb{R}^{d\times d}$~is the transformation matrix for each surrounding
context position, $b_r\in\mathbb{R}^d$~are the representation biases, and
$b_f\in\mathbb{R}$ is a bias for each word~$f\in V_F$.

The translation probability is given by
 $
\prob{f\given e_i} = \frac1{Z_{e_i}}\exp\lpa -\ener{f,e_i}
\rpa, \nonumber
$ 
where $ Z_{e_i} = \sum_{f} \exp\lpa-\ener{f,e_i}\rpa$ is the normalizer.

In addition to these translation probabilities, we

\noindent
have parameterized the
translation probabilities for the null word using a softmax over an additional
weight vector.

\subsection{Class Factorization}

We improve training performance using a class factorization strategy
\cite{Morin+al-2005}
as follows.
We augment the translation probability to be
$\prob{f\given e} = \prob{c_f\given e}\prob{f\given c_f, e}$
where $c_f$ is a unique predetermined class of~$f$; the class probability
is modeled using a similar log-bilinear model as above, but instead of predicting
a word representation~$r_f$ we
predict the class representation~$r_{c_f}$
(which is learned with the model) and we add respective new context matrices and biases.
Note that the probability of the word~$f$ depends on \emph{both} the class and
the given context words:  it is normalized only over words in the class $c_f$.

In our training we create classes based on word frequencies in the corpus as follows.
Considering words in the order of their decreasing frequency,
we add word types into a class until the total frequency of the word types
in the currently considered class is less than $\frac{\text{total tokens}}{\sqrt{\left|V_F\right|}}$
and the class size is less than~$\sqrt{\left|V_F\right|}$.
 We have found that the maximal class
 size affects the speed the most.

\section{Learning}

The original \fastalign{} model  optimizes the likelihood using the expectation
maximization (EM) algorithm where, in the M-step, the parameter update is
analytically solvable, except for the $\lambda$ parameter (the diagonal
  tension), which is optimized using gradient descent~\cite{naacl2013:dyer}.  We
modified the implementations provided with
\textsc{cdec}~\cite{dyer-EtAl:2010:Demos}, 
retaining its default parameters.

In our model, \fastalignLBL, we optimize the likelihood 
using the EM as well.
However, while training we fix the counts of the E-step to those computed
by \fastalign{}, 
trained for the default 5~iterations,
to aid the convergence rate, and optimize the M-step only.
Let $\theta$ be the parameters for our model. Then the gradient for each
sentence is given by
\vspace{-0.5em}
\begin{align}
\frac\partial{\partial\theta}
    \log&\,
          \prob{\svar fs \given \svar es} = \nonumber \\
    \sum_{k=1}^J&\sum_{l=0}^I \lsq
      \frac {            \prob{l\given k,I,J} \prob{\evar fsk \given \evar esl}}
            {\sum_{i=0}^I\prob{i\given k,I,J} \prob{\evar fsk \given \evar esi}}
            \right.\nonumber
            \\
   &
     \left.\phantom{\sum_{l=0}^I}
     \cdot
      \frac\partial{\partial\theta}
          \log\!\lpa\prob{l\given k,I,J} \prob{\evar fsk \given \evar esl}\rpa
          \rsq \nonumber
\vspace{-0.5em}
\end{align}
where the first part are the counts from the \fastalign{} model and second part comes from our model.

We compute the gradient for the alignment probabilities in the same way as in
the \fastalign{} model, and the gradient for the translation
probabilities using back-propagation~\cite{1986Natur.323..533R}.
For parameter update, we use \adagrad\ as the gradient descent algorithm
\cite{Duchi:2011:ASM:1953048.2021068}.

\section{Experiments}
\label{sec:experiments}

We first evaluate the alignment error rate of our approach, which establishes
the model's ability to both learn alignments as well as word representations
that explain these alignments.
Next, we use a cross-lingual document classification task to verify that the
representations are semantically useful. We also inspect the embedding space
qualitatively to get some insight into the learned structure.

\subsection{Alignment Evaluation}

We compare the alignments learned here with those of the \textsc{FastAlign}
model which produces very good alignments and translation \BLEU{} scores.  We
use the same language pairs and datasets as in~\newcite{naacl2013:dyer}, that is
the FBIS Chinese-English corpus, and the French-English section of the Europarl
corpus~\cite{koehn2005epc}.
We used the preprocessing tools from \textsc{cdec} and further replaced all
unique tokens with UNK. We trained our models with
100~dimensional representations for up to 40~iterations, and the \fastalign{}
model for 5~iterations as is the default.

Table~\ref{tab:aer} shows that our model learns alignments on part with those of
the \fastalign{} model. This is in line with expectation as our model was
trained using the \fastalign{}
expectations. However, it confirms that the learned word representations are
able to explain translation probabilities.
Surprisingly, context seems to have little impact on the alignment error,
suggesting that the model receives sufficient information from the aligned words
themselves.

\begin{table}
  \centering

\newcommand{\lgi}[2]{\textsc{#1}\textbar\textsc{#2}}
  \centering
  \begin{tabular}{lccc}
    \toprule
    Languages & \multicolumn{3}{c}{Model} \\
    \cmidrule{2-4}
    & \fastalign & \fastalignLBL & \fastalignLBL \\
    &            &   $k=0$ & $k=3$ \\
    \midrule
    \lgi{zh}{en} & 49.4 & 48.4 & 48.7 \\
    \lgi{en}{zh} & 44.9 & 45.3 & 45.9 \\
    \lgi{fr}{en} & 17.1 & 17.2 & 17.0 \\
    \lgi{en}{fr} & 16.6 & 16.3 & 16.1\\
    \bottomrule
  \end{tabular}
  \caption{\label{tab:aer} Alignment error rate~(\AER{}) comparison, in both
    directions, between the \textsc{FastAlign}~(\fastalign) alignment model and
    our model~(\fastalignLBL) with $k$~context words (see
      Equation~\ref{eq:energy}). Lower numbers indicate better performance.}

\vspace{-1em}
\end{table}

\subsection{Document Classification}

A standard task for evaluating cross-lingual word representations is document
classification where training is performed in one and evaluation in another
language. This tasks require semantically plausible embeddings (for
  classification) which are valid across two languages (for the semantic
  transfer).
Hence this task requires more of the word embeddings than the previous task.

We mainly follow the setup of~\newcite{klementiev-titov-bhattarai:2012:PAPERS}
and use the German-English parallel corpus of the European Parliament
proceedings to train the word representations. We perform the classification
task on the Reuters RCV1/2 corpus. Unlike
\newcite{klementiev-titov-bhattarai:2012:PAPERS}, we do not use that
corpus during the representation learning phase.
We remove all words occurring less than five times in the data and
learn 40~dimensional word embeddings in line with prior work.

To train a classifier on English data and test it on German documents we first project word
representations from English into German: we select the most probable German
word according to the learned translation probabilities, and then compute
document representations by averaging the word representations in each document.
We use these projected representations for training and subsequently test using
the original German data and representations. We use an averaged perceptron
classifier as in prior work, with the number of epochs (3) tuned on a subset of the
training set.

\begin{table}
  \centering
  
\begin{tabular}{lcc}
\toprule
Model  &en $\to$ de &  de $\to$ en  \\
\midrule
Majority class           &    46.8       &     46.8           \\
Glossed                  &    65.1       &     68.6         \\
MT                       &    68.1       &     67.4         \\
Klementiev et al.        &    77.6       &       71.1              \\
           BiCVM \textsc{Add} & {\bf 83.7} & 71.4 \\
           BiCVM \textsc{Bi} & 83.4 & 69.2 \\
\midrule
\fastalignLBL{} ($k=0$) &       82.8         & {\bf 76.0}           \\
\fastalignLBL{} ($k=3$) &      83.1         &  75.4          \\
\bottomrule
\end{tabular}
\caption{\label{tab:doccls} Document classification accuracy when trained on
  1,000 training examples of the RCV1/2 corpus (train$\to$test).
  Baselines are the majority class, glossed, and MT
  \cite{klementiev-titov-bhattarai:2012:PAPERS}.
  Further, we are comparing to
  \newcite{klementiev-titov-bhattarai:2012:PAPERS},
  BiCVM \textsc{Add} \cite{Hermann:2014:ICLR},
and BiCVM \textsc{Bi} \cite{Hermann:2014:ACLphil}.
  $k$~is the context size, see Equation~\ref{eq:energy}.
  }

\end{table}

Table~\ref{tab:doccls} shows baselines from previous work and classification
accuracies.
Our model outperforms the model
by~\newcite{klementiev-titov-bhattarai:2012:PAPERS}, and it also outperforms the
most comparable models by~\newcite{Hermann:2014:ACLphil} when training on German data
and performs on par with it when training on English data.\footnote{From
  \newcite[2014b]{Hermann:2014:ICLR} we only compare with models equivalent with
  respect to embedding dimensionality and training data. They still achieve the
  state of the art when using additional training data.}
It seems that our model learns more informative representations towards
document classification, even without additional monolingual language models or
context information. Again the impact of context is inconclusive.

\subsection{Representation Visualization}
\label{sec:visual}

\begin{figure}[t]
    \centering
  \vspace{-1em}
  \includegraphics[width=0.90\columnwidth,clip=true,trim=15mm 40mm 10mm 5mm]{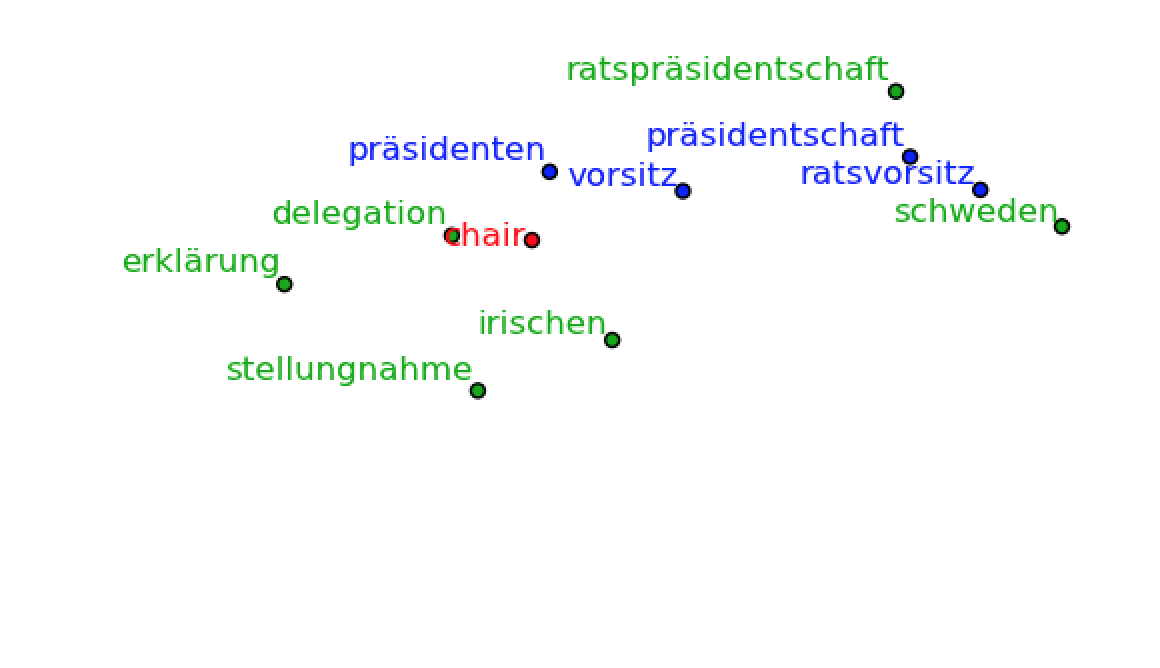}
  \vspace{-1em}
  \caption{A visualization of the expected representation of the
    translated English word \textit{chair} among the nearest German words:
words never aligned (green), and those seen aligned (blue) with it.
\label{fig:chair}}
\end{figure}

\begin{figure}[t]
  \centering
    \includegraphics[width=0.80\columnwidth,clip=true,trim=0 5mm 0 10mm]{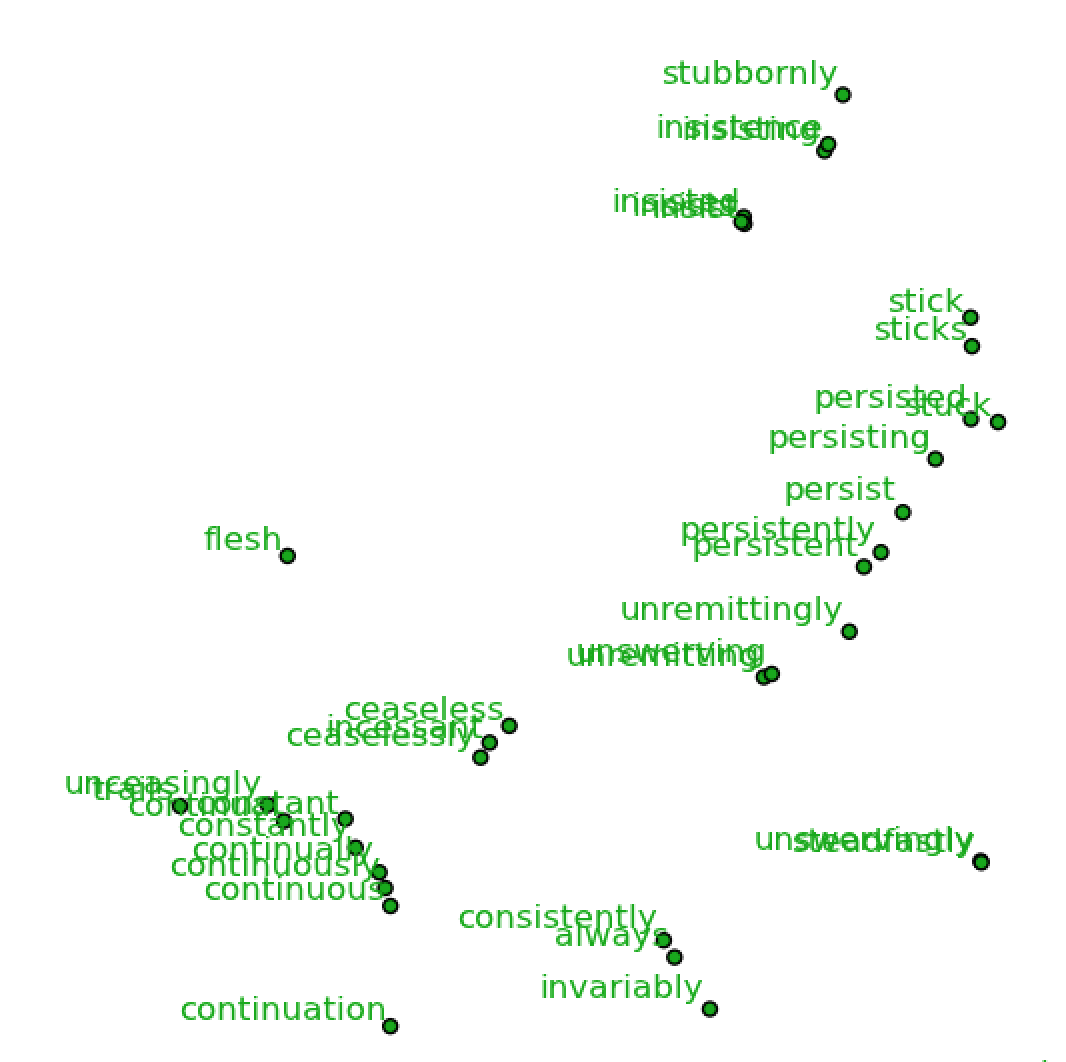}
  \vspace{-1em}
  \caption{A cluster of English words from the 10,000 most frequent English
    words visualized using t-SNE. Word representations were optimized
    for~$\prob{\text{zh}\given\text{en}}$ ($k=0$)\label{fig:mono1}.}
  \vspace{-1em}
\end{figure}

Following the document classification task we want to gain further insight into
the types of features our embeddings learn. For this we visualize
word representations using t-SNE projections \cite{Maaten:2008}.
Figure~\ref{fig:chair} shows an extract from our projection of the 2,000 most
frequent German words, together with an expected representation of a
translated
English word given translation probabilities.
Here, it is interesting to see that the model is able to learn related
representations for words \textit{chair} and \textit{ratspr\"asidentschaft}
(presidency) even though these words were not aligned by our model.
Figure~\ref{fig:mono1} shows an extract from the visualization of the 10,000
most frequent English words trained on another corpus. Here again, it is evident
that the embeddings are semantically plausible with similar words being closely
aligned.

\section{Conclusion}

We presented a new probabilistic model for learning bilingual word
representations. This distributed word alignment model (\model) learns both
representations and alignments at the same time. We have shown that the \model
model is able to learn alignments on par with the
\textsc{FastAlign} alignment model which produces very good alignments,
thereby determining the efficacy of the
learned representations which are used to calculate word translation
probabilities for the alignment task.
Subsequently, we have demonstrated that our model can effectively be used to
project documents from one language to another. The word representations our
model learns as part of the alignment process are semantically plausible and
useful. We highlighted this by applying these embeddings to a cross-lingual
document classification task where we outperform prior work, achieve results on
par with the current state of the art and provide new state-of-the-art results
on one of the tasks.
Having provided a probabilistic account of word representations across multiple
languages, future work will focus on applying this model to machine translation
and related tasks, for which previous approaches of learning such embeddings are
less suited. Another avenue for further study is to combine
this method with monolingual language models, particularly in the context of
semantic transfer into resource-poor languages.

\section*{Acknowledgements}
This work was supported by a Xerox Foundation Award and EPSRC grant number
EP/K036580/1. We acknowledge the use of the Oxford ARC.

\bibliographystyle{acl}
\bibliography{main}

\end{document}